**Development of a Voice-Controlled Tendon-Driven Bionic Hand**

Urja Kohli[1, a], Shagata Chanda[1, b], Kritika Gandhi[1, c], Charu Nigam[1, d], Aditi Surya Kamal[1, e], Pooja Bhati[1, f]

[1] Department of Mechanical and Automation Engineering, Indira Gandhi Delhi Technical University for Women, Delhi, India

[a] urja090btmae22@igdtuw.ac.in, [b] shagata063btmae22@igdtuw.ac.in,

[c] kritika012btdmam22@igdtuw.ac.in, [d] charu023btmae22@igdtuw.ac.in

[e] aditi051phd24@igdtuw.ac.in, [f] poojabhati@igdtuw.ac.in

## Abstract

The impairment of the hands can seriously affect the abilities of every individual to perform the every-day activity, so the design of stable and controllable support devices is a significant field of study. This paper is about the design and implementation of an automated bionic hand which is dedicated to the coordinated finger movement through the simplified and efficient actuation mechanism. The method that the proposed system was designed on is the tendon-based method whereby the servo motors generate the movement of the fingers, with assistance of the angular control which is calibrated. An actuation is controlled by a microcontroller that will be programmed by use of an Arduino-based microcontroller to carry out programmed gestures that include open hand, fist, pinch and half flexion. It has an interface that is voice command enabled to make it easy to interact with a Bluetooth based sender receiver architecture which offers an option of executing trained commands which are immediately converted to finger actions. To explore the motions behavior, finger coordination and control response to the input, the behavior of the experiment system is tested. The actuation of the fingers was found to take a total of about 7-8 seconds to achieve full flexion of all fingers in a sequence. The system showed repetitive and constant motion throughout several actuation cycles without loss of any apparent tension or precision of control. There was a stable grasp of objects of different shapes and sizes, which implied consistent coordination between the fingers. These findings indicate that the proposed system offers predictable and steady control behavior and has a simple and efficient mechanical and control architecture.

## Introduction

The amputation of one upper limb also greatly impacts the capacity of the person to execute the daily tasks like holding and manipulating objects [1]. One of the most significant needs of assistive technologies is the manufacture of prosthetic devices and orthoses, as the size of the segment of the global population requiring these devices is estimated at about 30 million individuals [2]. Moreover, the world statistics show that there are almost 10 million people with amputation of limbs, of which almost 3 million are victims of upper limb amputation [2], [3]. Although modern prosthetic and robotic hands have been developed, functional devices are still inaccessible to many people especially in developing countries like India where a huge percentage of the population still uses cosmetic or passive prostheses that offer a little functional assistance [3]. This shortcoming emphasizes the necessity of having assistive systems that should be efficient, dependable, and maneuverable and which should be applicable in a real-world environment. The human hand is characterized by a complicated anatomy and variability of its dimensions, which are essential in the design of prosthetics [4], [5]. Hand sizes vary in different populations and hence anthropometric data is important in maintaining the natural proportions of prosthetic devices to the human anatomy. Research on Indian people suggests that the average hand length is about 179 mm, and the average hand length of men is 186.2 mm, and 161.9 mm in women, and the mean breadth of the hands is about 82.2 mm [4], [6]. Moreover, according to the study made at Grant Medical College the mean lengths of fingers were 6.89 cm and 6.93 cm respectively among men and women [7]. These measurements present valuable reference information to be used to design prosthetic hands which have ergonomical fit of the hand and structural representation of the hand.

*Table 1 Review of Existing Prosthetic Hand Designs*

| **Author / Year** | **Key Contribution of Study** | **Method Used** | **Research Gaps** | **Relevance to Proposed System** |
|---|---|---|---|---|
| Srinivasan (2022) | Studied anthropometric measurements of human hands and identified differences in hand dimensions between individuals. | Anthropometric measurement of hand length, breadth and finger dimensions. | Provides statistical data but does not apply these measurements to prosthetic hand design. | Motivated the use of Indian anthropometric data to scale the prosthetic hand for improved ergonomic compatibility. |
| Sun et al. (2021) | Proposed a scientific method for determining morphological parameters for prosthetic hands. | Biomechanical modelling and morphological analysis. | Focused on morphology but did not address the affordability or accessibility of prosthetic devices. | Inspired by the need to combine morphological accuracy with low-cost design in prosthetic development. |
| Perez Romero et al. | Developed a sub-actuated anthropometric robotic prototype hand based on CT-derived hand geometry. | CT-based modelling and sub-actuated robotic mechanism. | Limited dataset for broader population generalization and lack cost-effective manufacturing considerations. | Inspired the idea of anthropometric scaling and simplified actuation mechanisms in prosthetic hand design. |
| Liarokapis et al. (2014) | Designed with a low-cost open-source prosthetic hand with compliant and modular under-actuated fingers. | Parametric anthropometric modeling and under-actuated finger mechanisms. | Primarily targeted partial hand amputations and limited integration of sensory feedback. | Inspired by the use of modular tendon-driven mechanisms and affordable prosthetic solutions. |

Different methods have been investigated to create anthropomorphic prosthetic hands that can balance between mechanical simplicity and the functional functionality [8]. Romero et al. built a sub-actuated robotic hand with CT based modelling to gain anatomically realistic geometry [9]. On the same note, Liarokapis et al. suggested a low-cost opensource prosthetic hand with compliant and under-actuated fingers to achieve flexibility among dissimilar users and degrees of amputation [10]. Additive manufacturing is a technique that has become common in the past few years in the fabrication of prosthetics because of its capability to generate complex geometries with great efficiency with minimal material consumption [11]. Polylactic Acid (PLA) and Acrylonitrile Butadiene Styrene (ABS) have become very common as a result of their power, resilience and easiness of shaping. In this case, especially, PLA is deemed biodegradable and can be used in biomedical fields because it is derived out of renewable materials [12].

Although this has been achieved, there are a number of constraints with the existing designs of the prosthetic hands. Most commercial systems are designed based on anthropometric data of the Western population, which might not be an accurate reflection of other end users in other parts of the world like India [4]. Besides this, it is still a major engineering problem to bring together actuators, sensors, control electronics, and power systems in a small and light-weight fabricated structure [3]. This is significant in solving these problems to achieve the creation of prosthetic hands that are functional, ergonomic, aesthetically acceptable, and affordable to a larger population.

The proposed study is dedicated to the design and development of an automatic bionic hand that incorporates anthropometric modelling, additive manufacturing, and the use of a microcontroller-based control. The hand structure is made up of 3D printing by use of the PLA material that allows it to be customized quickly. The design of the system is meant to offer a lightweight assistive device that can be utilized to help carry out simple grasping functions, with the future developments being the introduction of tactile feedback and the introduction of intelligent control measures. The suggested system is a voice-controlled bionic hand incorporating a voice recognition unit, Arduino control units, Bluetooth, and actuation by servo [16], [17], [18]. The voice recognition module processes spoken commands and sends them wirelessly to the receiver unit as command signals. The system has a set of predetermined gesture functions performed by controlled servo movement, which allows movement of the fingers flexion and extension in a coordinated way by a tendon-driven mechanism [14], [15].

The novelty of the proposed system is that it combines a low-cost tendon-actuated actuation system with a voice-controlled interface and allows the user to operate the system intuitively without a complex bio-signal acquisition system like EMG. In contrast to the traditional systems of prosthetic, which are based on costly sensors and complex control mechanisms, the proposed system is aimed at simplicity, low costs, and ease of implementation. The system shows a possible applicability in assistive applications like simple object manipulation, rehabilitation training and educational prototyping platforms, where the price and accessibility are paramount. Although the present prototype is controlled by pre-defined gestures, it offers a viable basis of more adaptive and user-friendly approaches in the development of prosthetics.

## 2. Methodology

### 2.1 System Control Architecture

The proposed bionic hand is a layered control system to translate human voice input into the actuation of the mechanical parts. The system is structured based on a pipeline with inputs, signal processing, commanding, and acting. The external data that is received in the system is a preset voice command, which is observed by a trained voice recognition module. The identified command is given a record number which is converted into a command ID by the microcontroller. This command ID is sent wirelessly and cross-linked to an existing gesture of the receiver end. The architecture maintains a deterministic association amid user input and finger movement, which can be depended upon to run gestures reliably and repeatably.

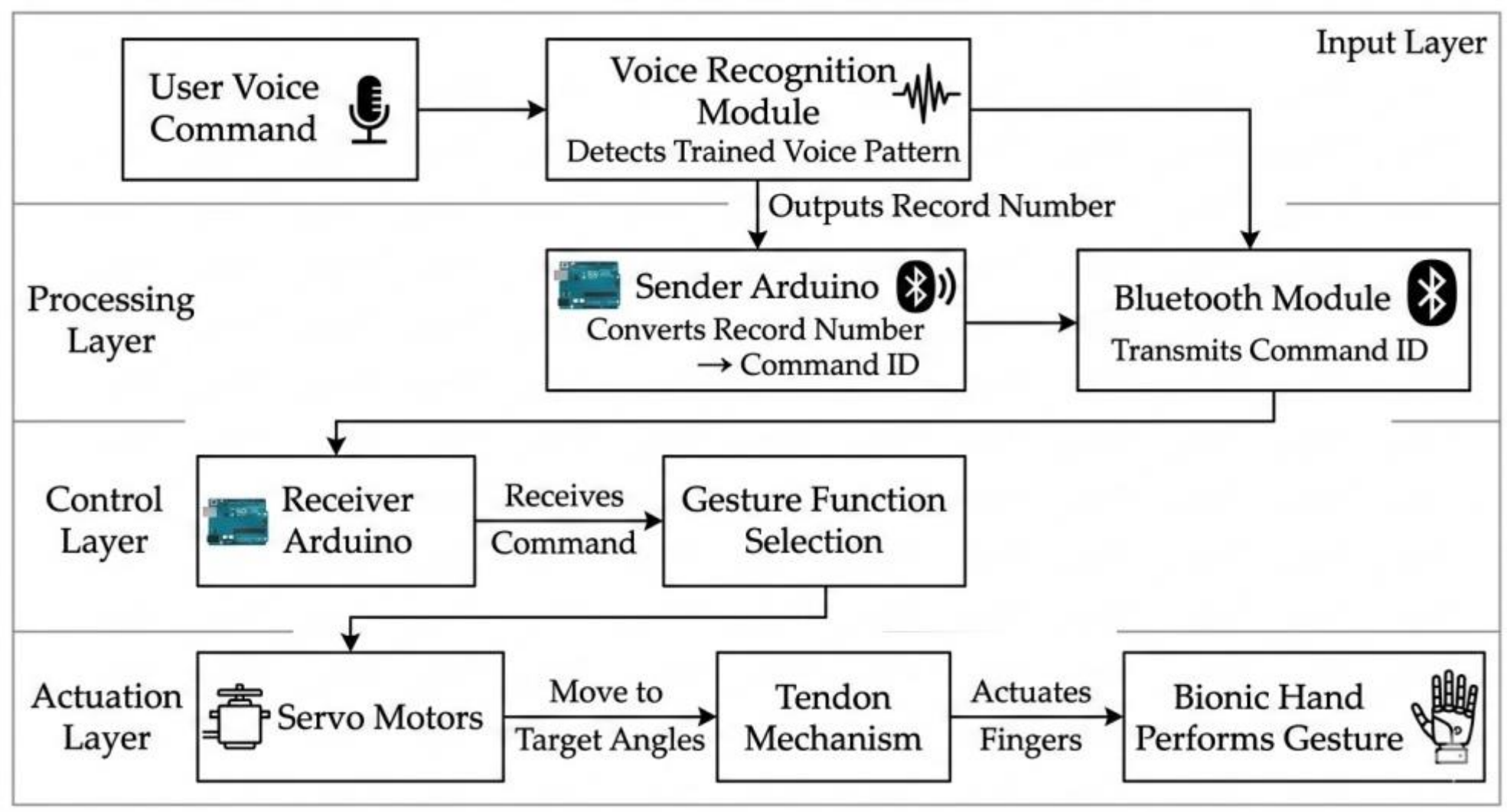

*Figure 1 Block diagram illustrating the architecture and operational flow of the proposed voice-controlled bionic hand system.*

**2.2 Voice Command Processing and Communication Pipeline**

The system is based on the sender-receiver structure to isolate voice recognition and actuation. The voice recognition module recognizes known voice patterns on the sender side and sends out a record number [17]. The microcontroller is fed with this record number which is then converted into a preset command ID. HC-05 Bluetooth units are set in a master-slave configuration and are used to establish wireless communication between the receiver and the sender [18]. The command ID is generated, and the sender sends it to the receiver using the serial communication. The microcontroller at the receiver end constantly monitors the received information and correlates the received command ID with its associated gesture action. This is initiated by a user uttering a trained command. The voice circuit recognizes the pattern and comes up with record number that is translated to command ID by the microcontroller. This command is sent through Bluetooth to the receiver where it is decoded and then applied to perform the equivalent gesture. This pipeline has guaranteed prompt communication with little delay when in operation.

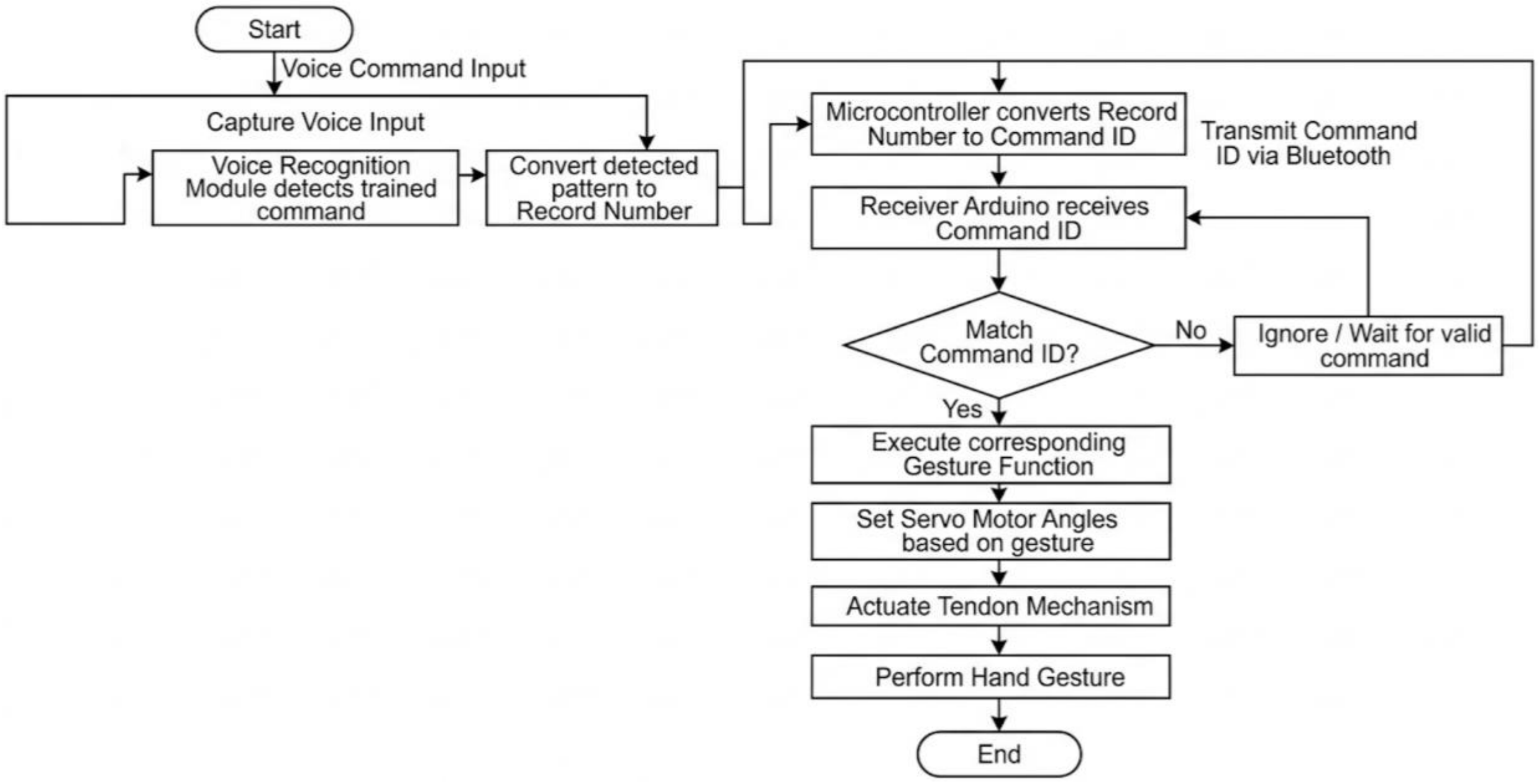

*Figure 2 Control flow of the voice-command processing and gesture execution in the bionic hand system.*

### 2.3 Control Logic and Gesture Mapping

The microcontroller provides a structured switch-case mechanism that provides the control logic of the system [16]. The command IDs are assigned to predefined functions which calculate the angular positions of the five servo motors that control the fingers [20]. The system allows movement of individual fingers as well as multi-finger movements since each finger is separately actuated. Gesture-level abstraction is implemented to make it easier to operate, which ensures that one command is issued to activate several actuators at once. The fixed gestures may be open hand, fist, pinch, grab, point, and relax. These gestures can be attained by allocating certain angular positions of each servo motor. The calibration of servo angles is done to depict various finger positions. A fully open position is approximately 90, a half-flexed position is approximately 45 and a fully closed position is approximately 0. Experimentally, these values were identified to have a smooth movement and at the same time not to cause an excess of the stress on the tendon mechanism.

*Table 2 Finger motion controlled through calibrated angular positions corresponding to biomechanical states.*

| Finger State | Servo Angle(°) (Approx.) | Functional Description |
|---|---|---|
| Fully Open | ~ 90 | Finger fully extended |
| Half Flexed | ~ 45 | Partial bending for grasp preparation |
| Fully Closed | ~ 0 | Complete flexion (fist/grip formation) |

The servo angle calibration was done by trial-and-error testing and observation of finger movement. First, each servo motor was rotated by a series of small steps, and the position of the fingers was determined visually. The chosen angles (approximately, 0, approximately 45, and approximately 90 degrees) were determined by attaining clear and repeatable states of the fingers in full closed, partially flexed, and fully open states. When calibrating, caution was taken to make sure that the tendon was not tightened excessively or left loose, which would avoid undue mechanical stress and loss of control [14]. There was also a tolerance of about ±5 degrees in attaining desired finger positions due to slight differences in the response of servos and tendon tension. Nonetheless, this difference did not have a significant impact on the general gesture performance, and the system was repeatable and reliable in terms of performance.

**2.4 Control System Framework**

The result of this is an open-loop control system, where a set of pre-defined input commands, which command actuator action directly, without any real-time feedback [19]. To create finger movement, it will consist of the voice command as the input, Arduino microcontroller as the controller and servo motors as the actuators. It is a method of simple and effective operation, in which it is ensured that the gestures that are preprogrammed are repeated each time. However, external changes to the system such as object resistance or grip force are not dynamically altered due to a lack of feedback. This was also experimentally found to be limited as the system was not capable of dynamically regulating the grip force and resulted in problems such as object slippage and inconsistent grip between objects with different properties.

**3. Implementation**

**3.1 Mechanical Design and Structural Development**

The mechanical structure of the bionic hand is designed to approximate the geometry and functional movement of a human hand using anthropometric references discussed earlier. The design adopts a multi-link articulated configuration, where each finger consists of segmented phalanges connected through rotational joints that enable controlled flexion and extension. This has resulted in a system capable of replicating the simple grasping and holding actions and keeping the structure of the system simple. Additive manufacturing is used to produce the components of the hand using Polylactic Acid (PLA), which was chosen due to its lightweight properties, easy production, and adequate strength to use as a prototype [11], [12]. 3D printing also allows quick prototyping and the ability to iterate by design and is especially useful in making prosthetic size more flexible. To achieve transmission of motion, a tendon-driven system is embraced by incorporation of nylon threads that run through the channels in the structure of the finger [14], [15]. When actuated, servo motors supply tension to these tendons causing the fingers to flex. As the tension is released, the re-establishing passive forces pull the fingers back to their original position. It is an easy to actuate method that is similar to biological tendons in operations. In the current implementation, a nylon monofilament fishing line was chosen as the tendon material due to its mechanical properties and availability. In order to implement the tendon mechanism, nylon monofilament fishing line (diameter 0.52 mm, breaking force 22 kg / 48.7 lb) was employed because of its high tensile strength, elasticity, and smooth surface that reduces friction in the finger channels. The chosen material makes sure that it is robust to withstand the forces of the MG996R high-torque servo motors in the system. These motors are operated at a constant torque with a 5V, 5A supply being regulated to operate under multi-finger actuation. It was demonstrated that the tendon system was consistent and reliable in performance with regard to repeated actuation cycles. The tendons were easily moved without any visible slack, breakage or loss of tension, even when all five servos were switched in parallel. The peak force absorbed by the motors was in the safe operating range of the tendon material, meaning that there was a suitable matching between actuator capability and tendon strength. However, fatigue behavior and long-term durability was not conducted in a systematic fashion in this study. Considering the long-term use, wear or minor elongation or changes in tension distribution may occur and would require regular inspections or replacement with the tendon.

### 3.2 Actuation Mechanism and Servo Integration

The bionic hand is actuated using five free-standing servo motors that drive the movement of each of the fingers. These servo motors transform electrical control messages into angular displacement that is very precise and sent to the fingers via the tendon system [20]. The angular positions of the servos are experimentally tuned to symbolize various finger states which include fully open and partially flexed and fully closed posture. Such a calibration is necessary to achieve a smooth and controlled movement without undue tension in the tendon mechanism, which otherwise may cause mechanical wear or upscale of the functioning of the system. Independent actuators are used to achieve the individual finger movement and also the multi-finger gestures. This is flexible and allows performing different predefined gestures, being consistent and repeatable in motion. The connection between the servo motors, microcontroller pins and the mechanical finger actuation are mapped in such a way to provide fine and independent control of each finger. Each servo is wired to a specific pin with a dedicated Pulse width modulation (PWM)-enabled pin and allows synchronized multi-finger movements to be controlled by programmed control signals [19].

*Table 3 Mapping of servo motor pins to fingers and mechanical actuation setting.*

| Finger | Servo Motor | Arduino Pin | Function |
| --- | --- | --- | --- |
| Thumb | MG996R | D5 | Opposition & grip support |
| Index | MG996R | D6 | Precision pinch control |
| Middle | MG996R | D9 | Central grip stability |
| Ring | MG996R | D10 | Supportive grasping |
| Little | MG996R | D11 | Grip closure completion |

The unit of the sender is made up of a voice recognition unit connected to an Arduino microcontroller, and a Bluetooth HC-05 module set to the master [13]. Voice recognition module identifies specific voice commands and displays a relevant record number which is handled by microcontroller and encoded into command identifier to be transmitted. The receiver unit consists of an Arduino microcontroller and a HC-05 Bluetooth unit that is set as a slave [18]. It is constantly analyzing the data that is sent by the sender and matches the identifiers of received commands to a set of gestures. The separation of the actuation and receiver units into modules makes the system more flexible, easier to debug, and enables the development of the input subsystem and actuation subsystems independently. Figure 3 depicts the practical implementation of the sender unit which consists of voice recognition unit and Bluetooth interface. It is designed on a breadboard in order to support comfortable prototyping and debugging of circuits between the microcontroller, communications module and input system.

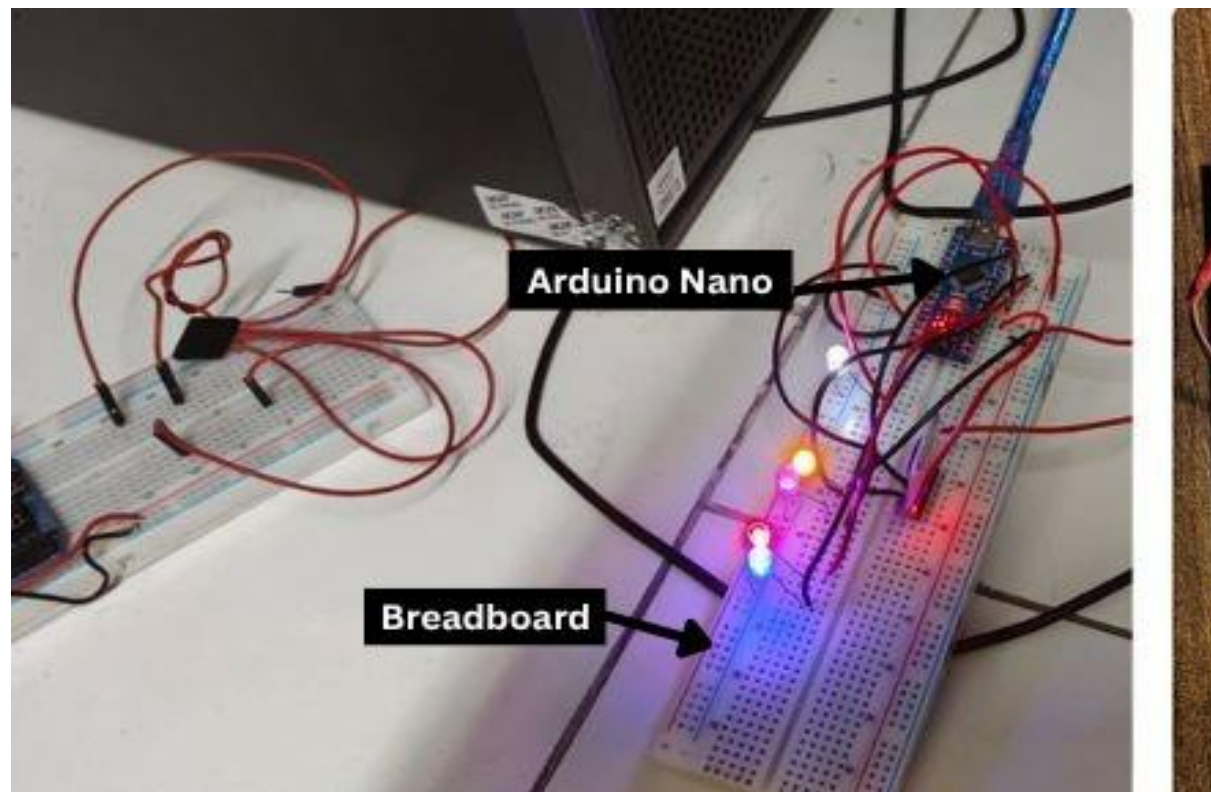

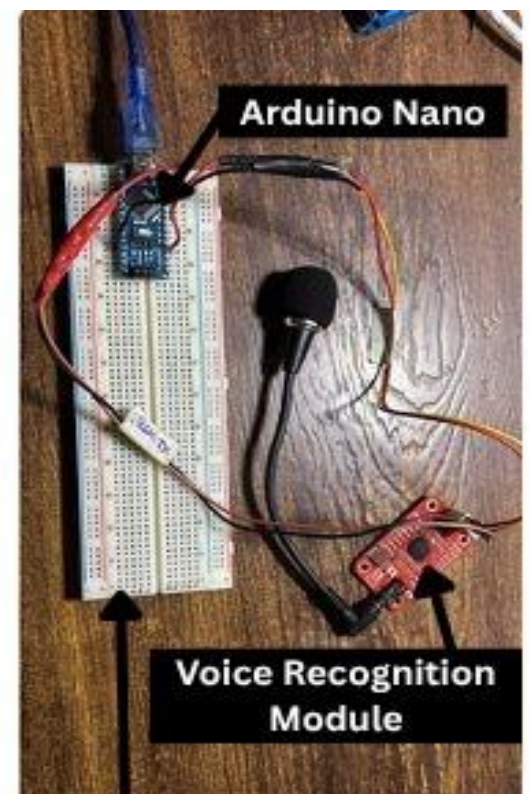

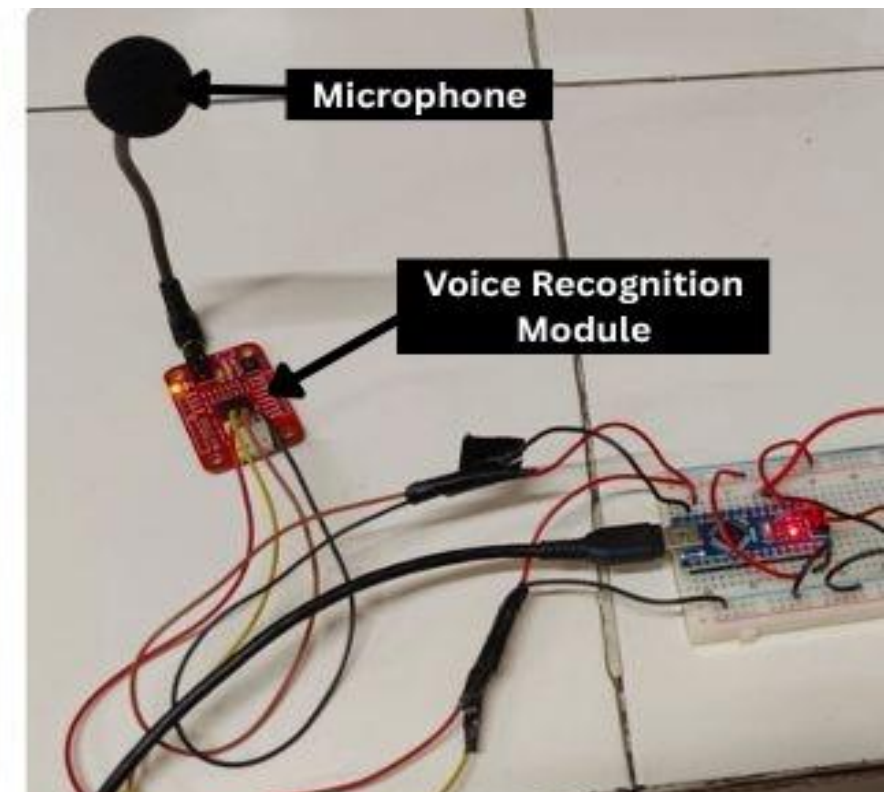

*Figure 3 Experimental design of the sender unit including voice recognition module, Arduino microcontroller and the Bluetooth interface.*

The Bluetooth modules are programmed on the AT command to put up a master slave connection. One of the modules is configured as the master device and the other module is positioned as a slave. The LED indicator on the HC-05 module has different blinking patterns during the process of pairing. First, the LED flashes at a high rate, which means that the module is at the pairing or discovery mode. After the successful connection between the master and slave modules has been made, the rate of blinking slows down a lot or it becomes constant which means that there has been a stable connection. This visual feedback is applied to confirm successful pairing prior to the commencement of transmission of data.

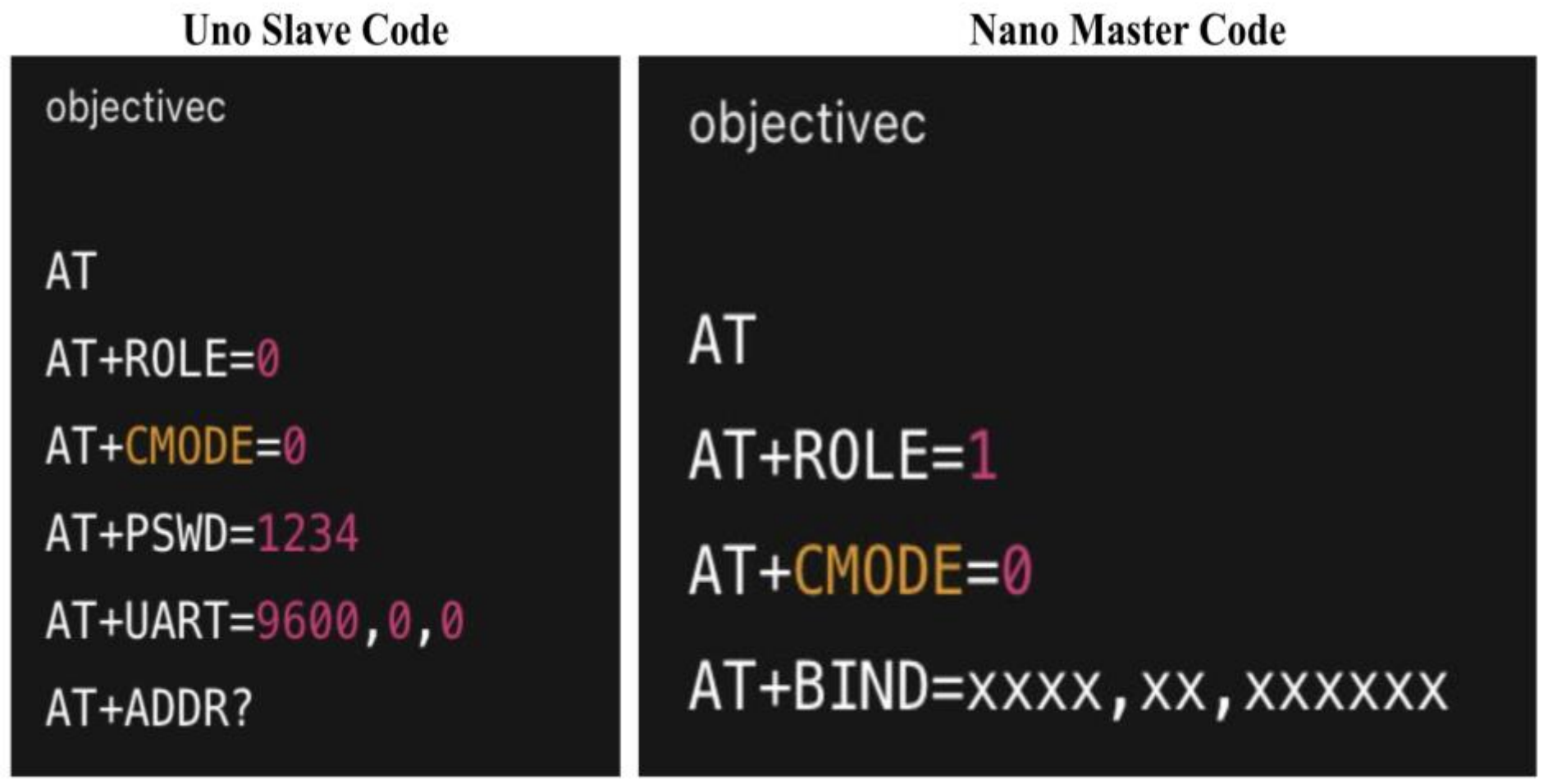

*Figure 4 AT command setup to use the HC-05 Bluetooth modules in master-slave mode*

**3.4 Circuit Design and Power Management**

The circuit design is such that voice recognition module, microcontrollers, Bluetooth modules, and servo motors are connected in the circuit and are made in one system. Both the voice module and Bluetooth modules are interfaced with the Arduino microcontrollers using serial communication protocols which provide reliability in relaying data. The servo motors get connected to the PWM enabled output pins of the receiver microcontroller which enables the angular displacement to be controlled with precise control. There is a standard grounding scheme, which is kept across all the components to provide a stable signal and reduce noise.

Power management is also of particular attention because servo motors consume more current when it is running [20]. Proper voltage control and power distribution is done to avoid fluctuation and maintain a stable operation of the system. There are proper practices of wiring so that the system can remain reliable and also the possibility of electrical interference is minimized. The microcontroller is the Arduino which forms the central control unit, producing PWM signals to control the angular position of each servo motor. PWM signals produced by the Arduino microcontroller are used to drive the servo motors. The frequency of the PWM is around 50 Hz, which is appropriate in the control of hobby servo motors. The angular position of each servo is calculated by changing the pulse width of the signal, generally between about 1ms (minimum position 0°) and 2ms (maximum position 180°) with values in between creating proportional angular movements. Arduino has adequate time resolution to provide controlled finger movements that are smooth and stable.

Since several high-torque servo motors will be used, an external power supply is delivered, which will stabilize the working process and will exclude the possibility of voltage drops that might influence the functioning of the system. A converter module is included in the circuit to simplify wiring and provide an efficient way of distributing power. The module allows the servo power connections to be consolidated, and the independent signal lines provided by the microcontroller to the motor, to provide an accurate motor control. The system has better reliability and low electrical noise by the separation of power and control pathways. A servo motor is tied to a unique pin in the microcontroller, meaning that programmed gestures can be used to generate synchronized and coordinated movements of the fingers.

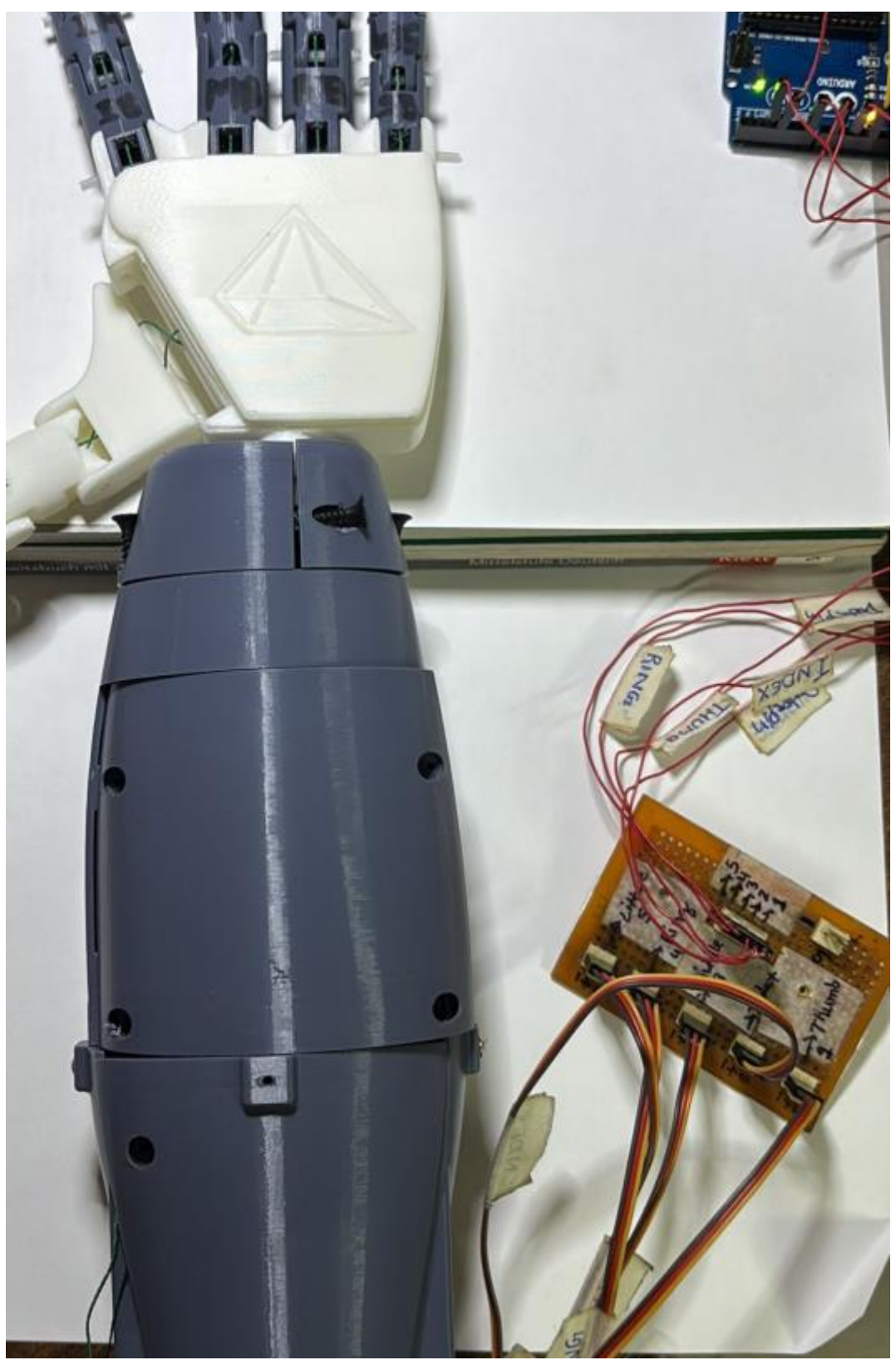

*Figure 5 Hardware system Integration of hardware components including servo driver connections, power distribution module and microcontroller interface.*

Figure 6 illustrates the circuit implementation in detail, showing how the voice module, Bluetooth module and the microcontroller are interconnected. The breadboard-based system has the flexibility of modification to the development and testing of the system.

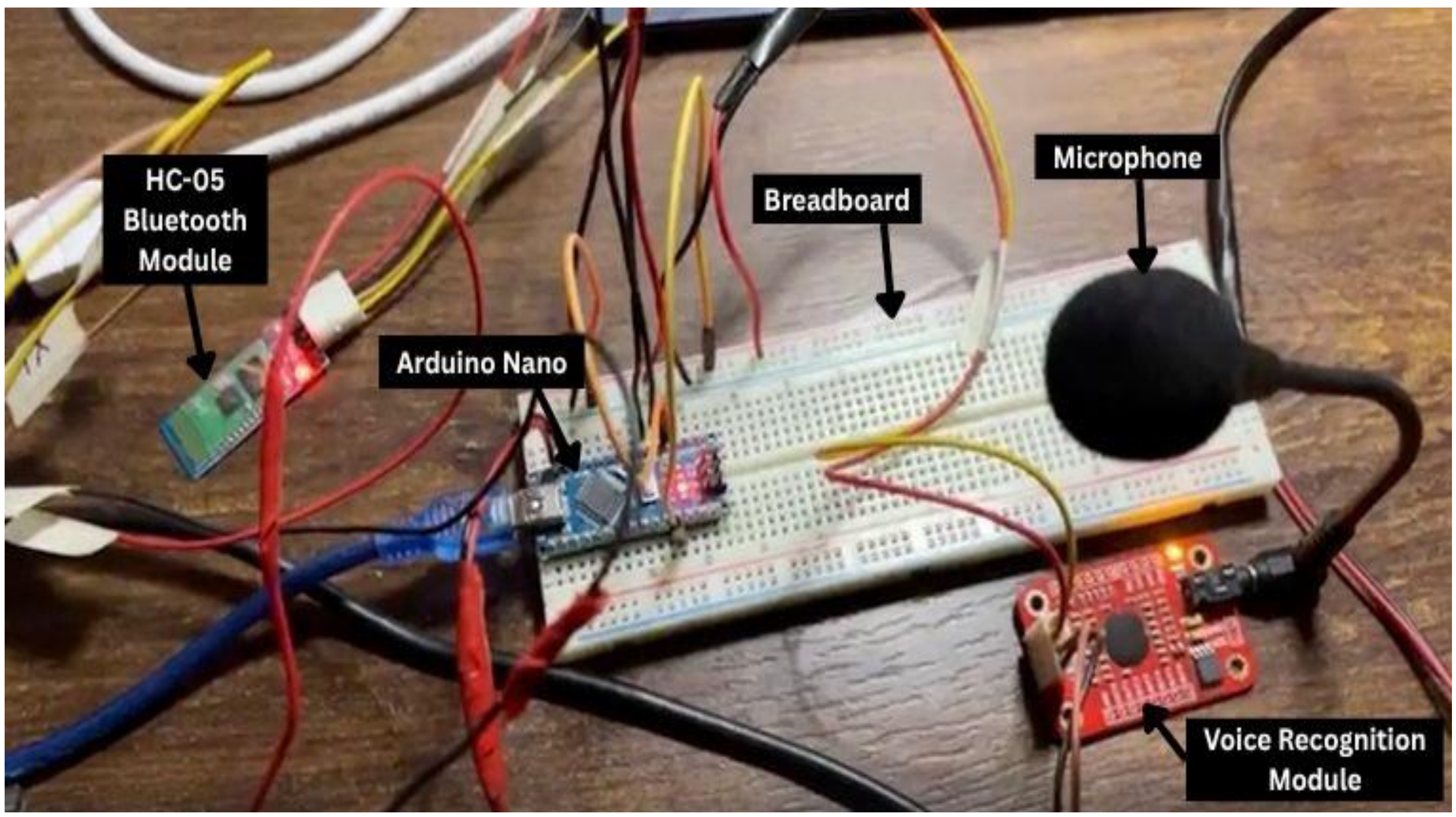

*Figure 6 Systematic circuit schematics of voice recognition and Bluetooth communication modules.*

**3.5 Embedded Software and Control Implementation**

The embedded C programming is used to implement the control system in the Arduino platform. The software has been designed so as to process input, communicate and control actuators in a sequential and deterministic way. On the sending end, the microcontroller interprets the output of the voice recognition module and matches the record number that is detected against a set command identifier. This is sent wirelessly through the Bluetooth unit. The microcontroller keeps on the receiver end constantly listens to incoming command data and feeds on it through a switch-case control structure. A command identifier is linked with a predetermined gesture operation, a gesture operation that assigns the angular orientation of the servo motors. Finger movements are coordinated by means of assigning each servo a particular angle. This structured programming model gives it a comfortable assurance that commands are being executed reliably as well as can be easily modified or extended to further utilize gestures.

*Table 4 Voice command to command ID mapping and gestures functionality.*

| Voice Command | Command ID | Function |
|---|---|---|
| Open | 12 | Opens all fingers |
| Close | 11 | Closes all fingers to form a fist |
| Grab | 14 | Semi-closed grip suitable for holding objects |
| Pinch | 13 | Thumb–index precision grip |
| Point | 15 | Index finger extended gesture |
| Relax | 16 | Neutral finger position |
| Thumb | 1 | Independent thumb movement |

### 3.6 System Assembly and Integration

The entire system is achieved by means of the incorporation of the mechanical framework, electronic parts and control software into one prototype. The palm is fitted with servo motors and the tendon cables are channeled through specific channels within each of the fingers that connect directly to the servo horns, so that flexion may be controlled. Special attention is also paid to the spatial location of electronic components so as not to disrupt mechanical motion. Servo actuation synchronous to finger joint axes is maintained critically to provide efficient transmission of the forces and minimization of the mechanical losses. The misalignment or asymmetry of tension of tendons may result in motion asymmetry or decreased grip performance. Integration process entails the iterative process of calibration especially in tendon tensioning and servo positioning which is required to provide smooth, coordinated, and repeatable motion of fingers. This refinement process will make the system run like a big machine and all fingers will act in unison.

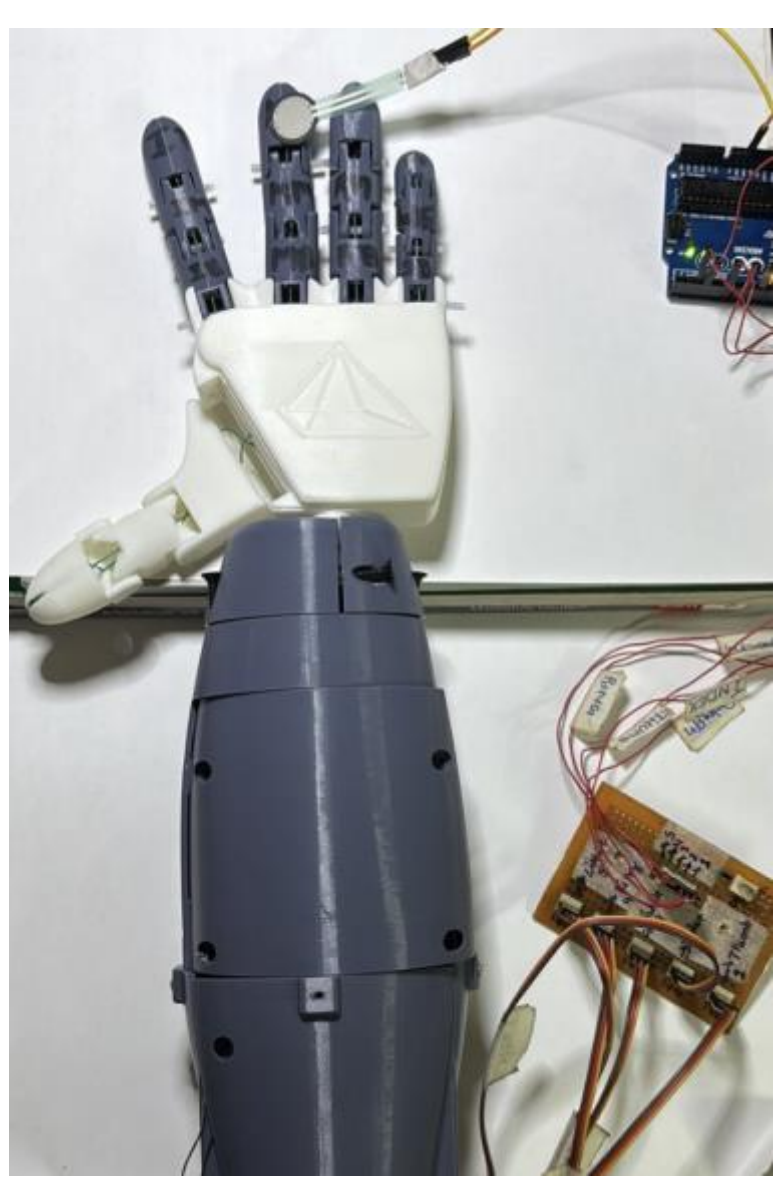

*Figure 7 Complete prototype of bionic hand assembled with servo mounted palm structure, tendon driven finger mechanism and built in electronic control system.*

**4. Results**

The functionality of the bionic hand developed was tested in order to measure the capacity of the bionic hand to execute controlled finger movement, produce synchronized actuation, and mimic the functions of the human hand. The system was shown to be reliable in terms of operation as the system operated well under its predetermined control inputs and the motion of the system was observed to be smooth and repeatable across numerous experiments. The association between servo actuation and finger movement was confirmed in terms of calibrated angular positioning, which is provided in Table 5. The state of each finger was mapped to an angle of one of the servos, making it possible to control how the finger transitions between fully open and partially flexed and fully closed. According to the findings, discrete angular control of servo motors can be used to achieve basic grasps poses (extension, or about 90 deg), intermediate flexion (or about 45 deg), and full closure (or about 0 deg). The system has its foundation on this mapping to predict and repeat finger motions.

*Table 5 Servo angle mapping of discrete finger states and functional positions of discrete finger states.*

| Finger State | Servo Angle (Approx.) | Functional Description |
|---|---|---|
| Fully Open | ~ 90 | Finger fully extended |
| Half Flexed | ~ 45 | Partial bending for grasp preparation |
| Fully Closed | ~ 0 | Complete flexion (fist/grip formation) |

The capability of the system to present functional gestures was also tested with regards to pre-defined gesture logic as presented in Table 6. The synchronous movements of several servo motors permitted performing open hand, fist, pinch and half-grip gestures. These gestures exhibit both uniform actuations, in which all the fingers act on the gesture at the same time, and differential actuation, in which certain fingers act separately in performing more fine tasks such as pinching. The findings validate the fact that the system is capable of both synchronized and selective movement, which is vital in the real-life uses of prosthetics.

*Table 6 Servo angle settings of preset gestures that allow different and synchronized movement of fingers.*

| Gesture | Thumb | Index | Middle | Ring | Little | Functional Use |
|---|---|---|---|---|---|---|
| Open Hand | 90 | 90 | 90 | 90 | 90 | Rest position |
| Fist | 0 | 0 | 0 | 0 | 0 | Power grasp |
| Pinch | 45 | 45 | 90 | 90 | 90 | Small object holding |
| Half Grip | 45 | 45 | 45 | 45 | 45 | Medium grasp |

The system performance is also further validated with visual observation of the finger movement during actuation as depicted in Figure 8. The series of photos depicts various points of finger flexion with the help of tendon action servo control. As the servo motors turn, the tendons are stretched, and joints of the fingers bend under control [14], [15].

The transition between extended and flexed forms exhibits continuous and smooth movement which is similar to fundamental finger kinematics of humans. The system shows the similarity in the behavior of various fingers, which validates the efficiency of the tendon transmission system in decoding rotational motor input into the coordinated joint motion. There is some slight difference in bending in different fingers because of the variation in routing of tendons and mechanical toleration but the overall movement is constant and coordinated.

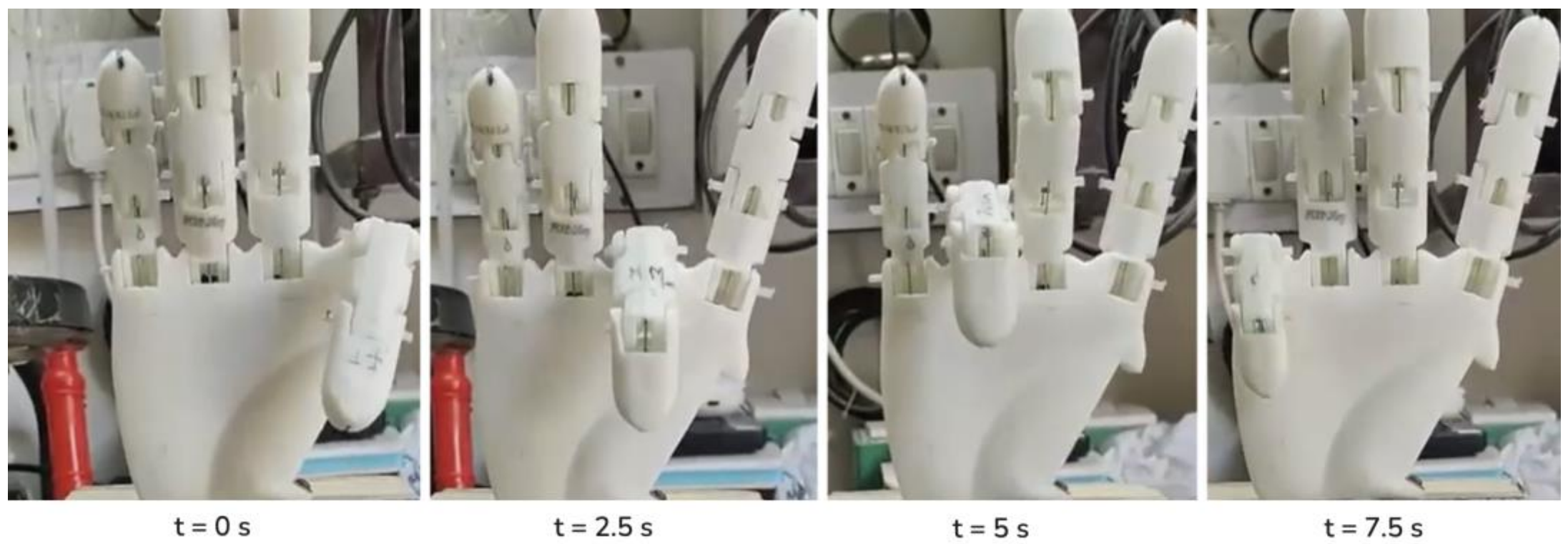

*Figure 8 Sequential finger flexion during tendon-driven actuation using servo motors, showing time progression of finger movement (t = 0s to 7.5s).*

Object interaction tests, which are presented in Figure 9, further illustrate the practical capability of the system. The bionic hand could also be used to hold objects of various shapes and sizes such as cylindrical shapes and irregular shapes. The fingers inherently change the way they are placed on the surface of the object during grasping. The grip force is not actively controlled, but the system can keep the objects stable enough to be manipulated easily. These observations prove that the suggested design is not restricted to a set of possible motions but can be applied to the real-life objects to experience a functional interaction. Although the system has shown consistency in grasping over a variety of items, some limitations, specific to the open-loop control system, were realized during experimental trials. There was minor slipping when dealing with smooth-surface objects because of the lack of real-time force control. Moreover, the thickness of objects affected gripping abilities, which led to either poor grip or too tight. The results show that the system cannot dynamically respond to changes in object resistance and surface properties. Testing objects were soft and rigid materials of different sizes to test the stability of grasp in different conditions. The size of the objects was approximated to be between 1 cm and 5 cm.

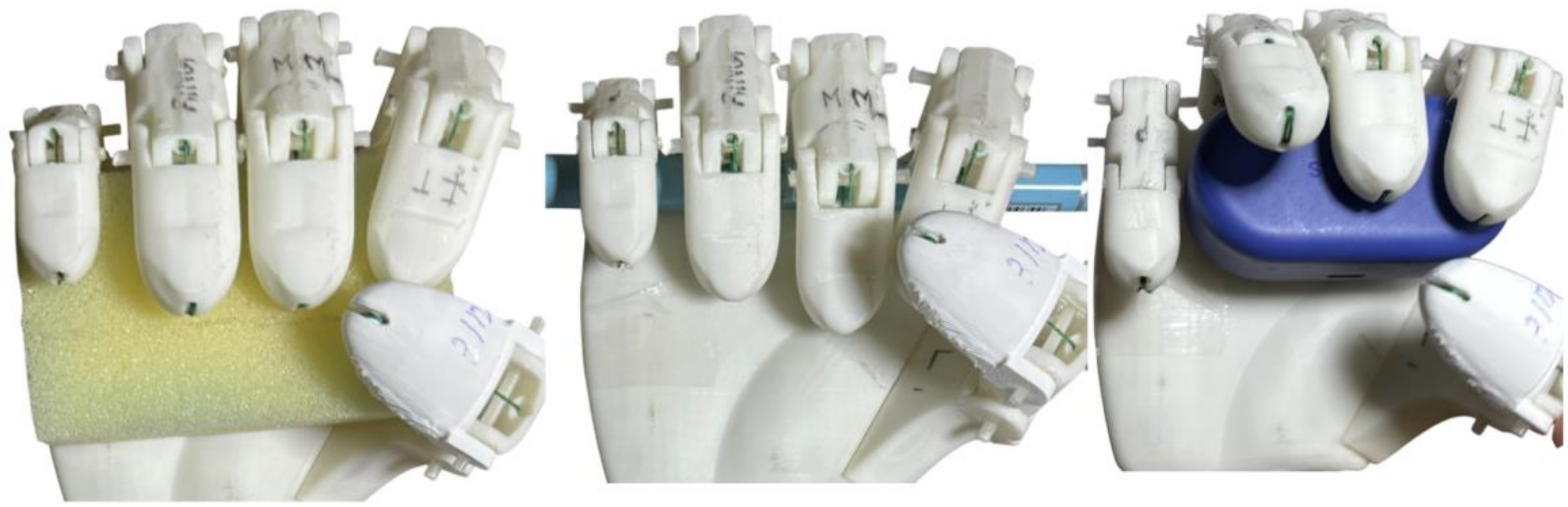

*Figure 9 Object grasping performance that involved coordinated finger movement and adaptability of the tendon-driven bionic hand across objects of varying shapes and sizes.*

This response is the dynamic response of the system to actuation of a finger, which is investigated by the trajectory of the angular position of joints through time, as in Figure 10. The drawn curves depict the variations in the angles of each finger in the flexion, which allows one to comprehend the behavior of the servo-controlled movement with time. Angular position is gradually and smoothly increased in all fingers, which is the sign of stable control and absence of sudden and irregular movement. The similarity in the curves indicates that several fingers move in a coordinated manner as per the control inputs. There are some differences in the speed of the individual fingers to bend, could be held back by the fact that tendons may differ in tension, positioning, and mechanical disparities in their structure.

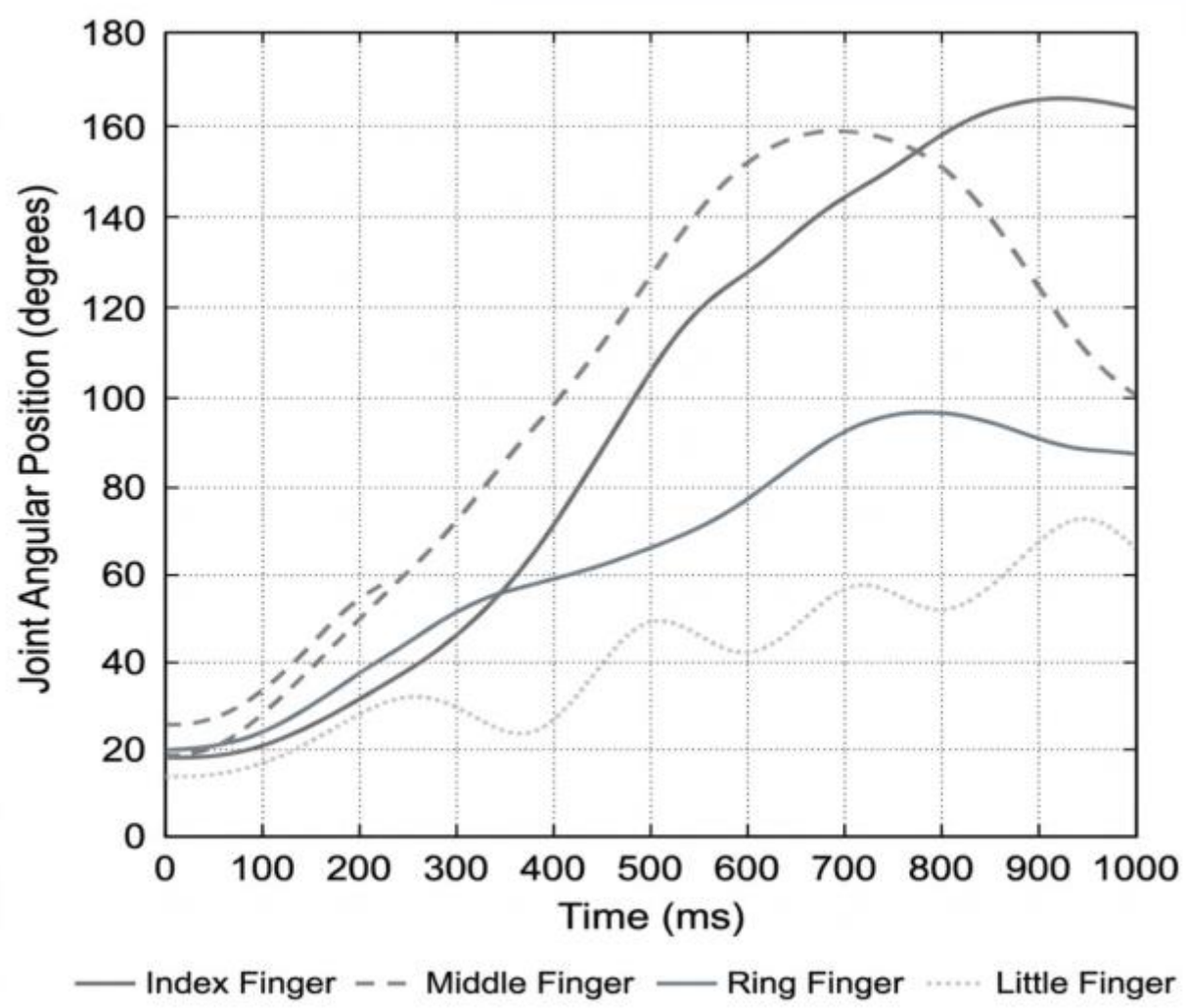

*Figure 10 Finger joint angular position (degrees) versus time (ms) during flexion using servo actuation for multiple fingers.*

The voice recognition module was tested in its performance by training and loading it with predefined control commands to the system [17]. The output in the serial monitor as in Figure 11 has proved that every command has been recorded and loaded.

**Code Commands**

```
switch (cmd) {
  // Your 7-command system
  case 12: openHand();   break;
  case 11: fist();       break;
  case 14: grab();       break;
  case 13: pinch();      break;
  case 15: point();      break;
  case 16: relaxHand();  break;
  case 1:  thumbOpen();  break;
```

**Successfully Loaded Commands**

```
Output    Serial Monitor ×
Message (Enter to send message to 'Arduino Uno'
Record 6        Trained
--------------------------------------
load 0 1 2 3 4 5 6
--------------------------------------
Load success: 7
Record 0        Loaded
Record 1        Loaded
Record 2        Loaded
Record 3        Loaded
Record 4        Loaded
Record 5        Loaded
Record 6        Loaded
```

*Figure 11 Serial monitor with good training and loading of voice commands to control the bionic hand.*

The finger movements were connected with each identified voice input and allowed one to open hand, fist, pinch and point. Voice recognition was also exhibited in the system with no observable delay, meaning that there was a good interaction between the voice module and the microcontroller. This proves that voice-based control could be successfully employed as an input device in controlling bionic hand. In general, repeatability and consistency of the motion patterns are obtained by the system; this validates the usefulness of PWM-based servo control in tendon-driven finger movement. Minor anomalies of certain curves represent actual mechanical constraints yet have no impact on the general operation of the system [19].

## Conclusion

This paper has introduced the design and development of an automated bionic hand that has a tendon-driven actuation system that can make simple finger movements. The system combines mechanical design, electronics and Arduino based controls to have a coordinated movement among several fingers. The tendon mechanism is successful in transforming the rotational movement of the servo motors into controlled flexion of the fingers to form motion which is very similar to the simple movement of the human hand. Calibrated servo angles are used to provide predefined movements, including the open hand, fist, pinch, and half flexion which are repeatable and constant. The experiment results demonstrate smooth and stable movement of fingers, which means good performance of the system. Voice-based control also allows an intuitive control where the trained commands are correctly identified and translated into finger movements. This shows that the system can react to simple external stimuli hence it can be used in assistive and robotic applications. Overall, the prototype offers a practical and modular method of controlled finger actuation that offers high repeatability and is simple to implement. While the existing testing is restricted to functional and mechanical performance testing as human user validation involves extra safety measures and ergonomic evaluation, future research will be based on the user-based evaluation to comprehend usability real-life better.

In order to improve the system further, the introduction of a closed-loop feedback approach with the integration of Force Sensitive Resistor (FSR) sensors is suggested. Such sensors may be strategically installed at the fingertips to sense the force of contact when interacting with objects. The values of the sensed forces can be constantly measured by the microcontroller and utilized in dynamically setting the positions of servo motors so that adaptive grip control can be achieved depending on the properties of the object, i.e. its shape, size, and surface texture. This can be done by a threshold-based control scheme where the servo actuation is adjusted once a set force limit has been attained, avoiding such problems as slippage or overgripping. This incorporation would enable the system to shift to a closed-loop system largely enhancing safety, accuracy, and practicality in the real world. The system may also be enhanced with the use of more control interfaces, like voice recognition, gesture control, or bio-signals (EMG), allowing more human-friendly and responsive human-machine interaction.

## Author contributions

U.K.: methodology, embedded software and control implementation, voice recognition and Bluetooth integration, system assembly and integration, tendon routing and servo calibration, experimental testing, data analysis, writing (original draft). S.C.: fabrication and post-processing of 3D-printed components, system assembly. K.G.: literature review. C.N.: fabrication and post-processing of 3D-printed components. A.S.K.: fabrication of 3D-printed components, writing (review and editing). P.B.: conceptualization, supervision, writing (review and editing).